\documentclass[9pt]{article}
\usepackage{spconf,amsmath,graphicx}
\usepackage{CJKutf8}
\usepackage{booktabs}
\usepackage{multirow}
\usepackage[ruled, linesnumbered]{algorithm2e}
\usepackage{subfigure}
\usepackage{xcolor}

\title{LEAPT: Learning Adaptive Prefix-to-prefix Translation For Simultaneous Machine Translation}
\name{Lei Lin, Shuangtao Li, Xiaodong Shi\thanks{Equal Contribution: Shuangtao Li. Corresponding author: Xiaodong Shi. This work is supported by National key R\&D Program of China (Grant no.2020AAA0107904), the Key Support Project of NSFC-Liaoning Joint Foundation (Grant no. U1908216). and the Major Scientific Research Project of the State Language Commission in the 13th Five-Year Plan (Grant no. WT135-38).}}
\address{Key Laboratory of Digital Protection and Intelligent Processing of Intangible Cultural Heritage
\\
of Fujian and Taiwan (Xiamen University), Ministry of Culture and Tourism, China}
\begin{document}
%\ninept
%
\maketitle

\begin{abstract}
\vspace{-0.5em}

Simultaneous machine translation, which aims at a real-time translation, is useful in many live scenarios but very challenging due to the trade-off between accuracy and latency. To achieve the balance for both, the model needs to wait for appropriate streaming text (READ policy) and then generates its translation (WRITE policy). However, WRITE policies of previous work either are specific to the method itself due to the end-to-end training or suffer from the input mismatch between training and decoding for the non-end-to-end training. Therefore, it is essential to learn a generic and better WRITE policy for simultaneous machine translation. Inspired by strategies utilized by human interpreters and ``wait'' policies, we propose a novel adaptive prefix-to-prefix training policy called \textbf{LEAPT}, which allows our machine translation model to learn how to translate source sentence prefixes and make use of the \emph{future} context. Experiments show that our proposed methods greatly outperform competitive baselines and achieve promising results.

% The abstract should appear at the top of the left-hand column of text, about
% 0.5 inch (12 mm) below the title area and no more than 3.125 inches (80 mm) in
% length.  Leave a 0.5 inch (12 mm) space between the end of the abstract and the
% beginning of the main text.  The abstract should contain about 100 to 150
% words, and should be identical to the abstract text submitted electronically
% along with the paper cover sheet.  All manuscripts must be in English, printed
% in black ink.
\end{abstract}
\vspace{-0.6em}
\begin{keywords}
Machine Translation, Simultaneous Machine Translation
\end{keywords}
\vspace{-1.8em}
\section{Introduction}
\label{sec:intro}
\vspace{-1em}

Simultaneous machine translation (SiMT) generates partial sentence translation (a target prefix) while receiving a streaming source input(a source prefix), thus demanding high translation quality and an as-short-as-possible delay. Compared with classic full-sentence machine translation, simultaneous machine translation is more challenging due to the incomplete source sentence during translating (see Figure~\ref{fig:fig1}).

A traditional streaming simultaneous translation (ST) system is usually formed by cascading a streaming auto-speech-recognition (ASR) component with a streaming machine translation (MT) component~\cite{oda2014optimizing,dalvi2018incremental}. Most of the previous work focuses on simultaneous text translation~\cite{gu2016learning,cho2016can,arivazhagan2019monotonic,ma2018stacl}, where main research direction is to figure out a reasonable READ policy and WRITE policy. Recent work generally splits into two classes:

\begin{figure}[!t]
    \centering
    \subfigure[Full-sentence NMT]{
        \includegraphics[width=2.55in]{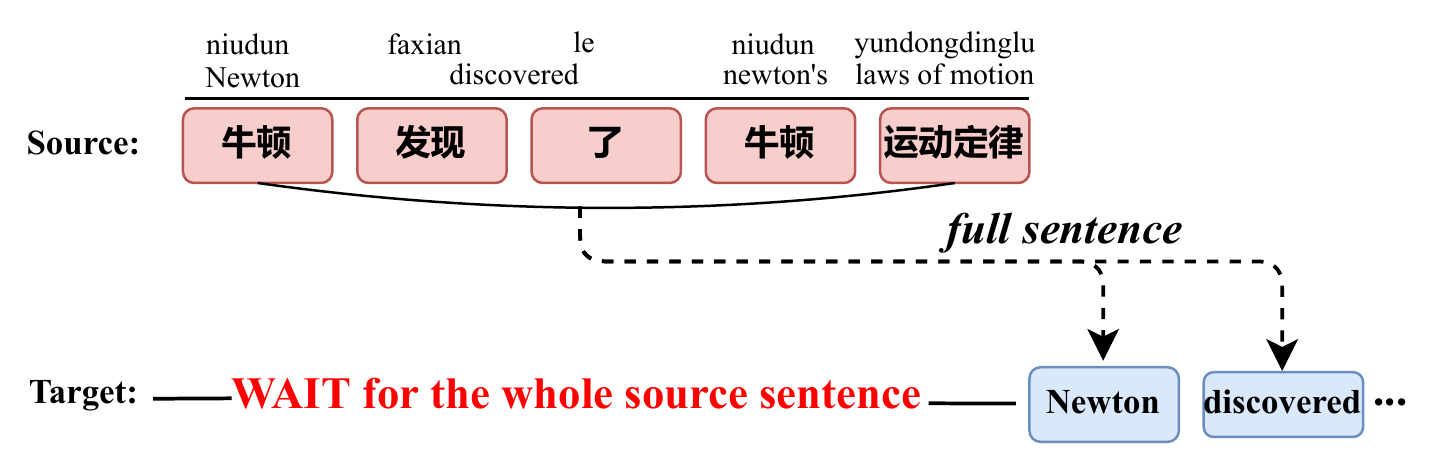}
    }
    \subfigure[SiMT]{
        \includegraphics[width=2.50in]{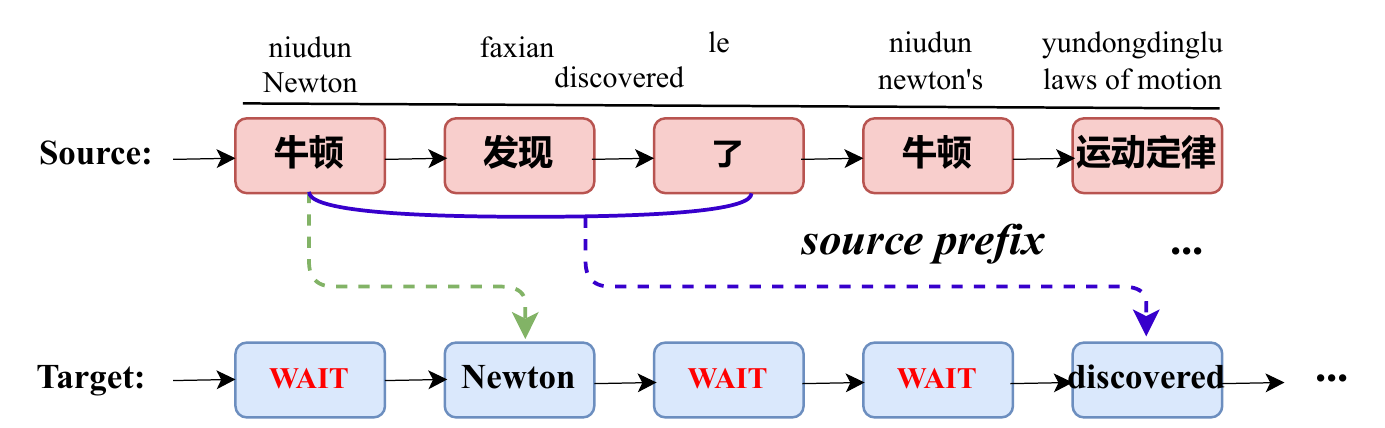}
    }
    \vspace{-1.0em}
    \caption{Overview of full-sentence NMT and SiMT.}
    \label{fig:fig1}
    \vspace{-1.5em}
\end{figure}

\vspace{-0.6em}
\begin{itemize}
    \item{\textbf{READ Policies}} determine whether to wait for another source word (READ). They can further fall into two classes commonly. One is fixed policies that segment the source sentence based on a fixed length~\cite{dalvi2018incremental,ma2018stacl,ma2020simulmt,zhang2021universal}. For example, the wait-\emph{k} method needs to read \emph{k} source words first, and then generates one target word continually after receiving each following source word. Although policies of this type are intuitive and simple to implement, they usually lead to a drop in translation quality owing to lacking consideration of contextual information. The other is adaptive policies that segment the source sentence according to dynamic contextual information. They either use a specific model~\cite{sridhar2013segmentation,cho2016can,gu2016learning,zheng2019simpler,zheng2020simultaneous,zhang2020learning,zhang2022learning} or end-to-end (e2e) approaches~\cite{arivazhagan2019monotonic,zheng2019simultaneous,ma2019monotonic} to chunk the streaming source text. The adaptive methods can form more flexible segments than the fixed ones and attain better translations.
    \vspace{-0.4em}
    \item{\textbf{WRITE Policies}} translate source sentence prefixes produced by READ policies. Previous work concentrates on non-end-to-end (ne2e) approaches~\cite{zhang2020learning,gu2016learning,cho2016can} or e2e approaches~\cite{ma2019monotonic,zhang2022modeling,ma2018stacl} to translate source sentence prefixes. For e2e approaches, they are specific to the method itself and may suffer from error propagation caused by wrong segmentation. For ne2e approaches, they do not consider the input mismatch between training and inference.
    % i.e., trained on the sentence pairs while decoding based on the source prefixes. 
    Therefore, it is important to find a generic and better WRITE policy for ne2e approaches.
    
    % However, they do not consider contextual information and consistency between partial prefix translations and usually result in a drop in translation quality even if READ policies generate flexible source sentence prefixes. Therefore, it is important to find a policy for learning adaptive prefix-to-prefix MT model.
    
    % a prefix-to-prefix MT model to translate the source sentence prefix. However, they do not perform well enough in translating the source sentence prefixes mainly due to the input mismatch between training and inferencing, where the full MT model is intuitive while the prefix MT model trained with fixed segmentation polices do not consider contextual information.
    
    % They can further fall into two classes commonly. One is full sentence MT model that translates the source sentence prefix, which is trained on parallel sentence pairs rather than sentence prefix pairs. For example, the ASP method segments the source sentence adaptively, and then translates the source sentence prefix using the full sentence MT model. Although they are more flexible in segmenting the source sentence, the subsequent full sentence MT model used for translating the source sentence prefix limits overall translation quality. The other is prefix-to-prefix MT model that appears as the opposite of the previous ones.
\end{itemize}
\vspace{-0.7em}

\begin{table}[!t]
    \centering
    \small
    % \renewcommand{\arraystretch}{1.4}
    % \resizebox{7.5cm}{!}{
    \resizebox{1.0\linewidth}{!}{
    \begin{tabular}{c|ccc}
    \hline
    \multirow{2}{*}{Sentence Pair} & \multicolumn{3}{c}{\begin{CJK*}{UTF8}{gbsn} 牛顿 \ 发现 \ 了 \ 牛顿 \ 运动定律 \end{CJK*}} \\
    \cline{2-4}
    & \multicolumn{3}{c}{Newton discovered newton's laws of motion} \\
    \hline
    \multirow{2}{*}{MU Pairs} & \multicolumn{1}{c|}{\begin{CJK*}{UTF8}{gbsn} 牛顿 \end{CJK*}} & \multicolumn{1}{c|}{\begin{CJK*}{UTF8}{gbsn} 发现 \ 了 \end{CJK*}} & \begin{CJK*}{UTF8}{gbsn} 牛顿 \ 运动定律 \end{CJK*} \\
    \cline{2-4}
    & \multicolumn{1}{c|}{Newton} & \multicolumn{1}{c|}{discovered} & newton's laws of motion \\
    \hline
    \end{tabular}
    }
    % }
    \vspace{-0.8em}
    \caption{The workflow of a simultaneous interpreter.}
    \label{table1}
    \vspace{-1.5em}
\end{table}
% When doing simultaneous translation, an interpreter first judges current received streaming text whether constitutes an MU and then translates them monotonically.
In this paper, we propose an adaptive prefix-to-prefix training policy for simultaneous machine translation. Our method is inspired by strategies utilized by human interpreters in simultaneous interpretation, who translate source sentence prefixes based on \emph{Meaningful Unit} (MU) with clear and definite meaning, which can be directly translated monotonically (see Table~\ref{table1}). Specifically, when listening to speakers, interpreters usually preemptively group the streaming words into MUs (READ policy), and then  perform translation monotonically while making the translation grammatically tolerable (WRITE policy). By doing so, interpreters can maintain high translation quality and an as-short-as-possible delay. Therefore, one natural question can be raised: can we train the WRITE policy to learn how to translate source sentence prefixes like human interpreters? To answer this question, we propose a novel translation-based method to generate pseudo prefix pairs, since there are no standard adaptive prefix pairs for training. Specifically, the method detects whether the translation of a source sentence prefix is a prefix of the full sentence's translation. If so, we regard a source sentence prefix and its translation as a prefix pair. Besides, we further propose a refined method to make use of the \emph{future} context to generate parallel prefix pairs with \emph{m} future words to improve the translation quality and consistency between partial prefix translations. Finally, the prefix-to-prefix MT model is jointly trained on the original sentence pairs and pseudo prefix pairs. 
% Our methods outperform previous WRITE policies, as they translate source sentence prefixes according to contextual information.

Experimental results on NEU 2017 Chinese-English (Zh-En) and WMT 2015 German-English (De-En) datasets show that our approaches outperform the strong baselines in terms of translation quality and latency. The contributions of this paper can be summarized as follows:
\vspace{-0.5em}
\begin{itemize}
    \item Inspired by strategies utilized by human interpreters, we propose a novel adaptive prefix-to-prefix training policy for SiMT, which produces high-quality translations with both low latency and high latency.
    \vspace{-0.6em}
    \item Inspired by ``wait'' policies, we further propose a refined method to make use of the source sentence prefixes' \emph{future} context by adding \emph{m} future words.
    \vspace{-0.6em}
    \item Our method is effective and simple, which achieves 19.6 (45\% relative improvement), and 27.2 (35\% relative improvement) BLEU scores for Zh-En and De-En experiments respectively in low latency compared with strong baseline model.
\end{itemize}

% However, there is a gap exits in the input of full sentence MT between training and inferencing. 

% These guidelines include complete descriptions of the fonts, spacing, and
% related information for producing your proceedings manuscripts. Please follow
% them and if you have any questions, direct them to Conference Management
% Services, Inc.: Phone +1-979-846-6800 or email
% to \\\texttt{icip2022@cmsworkshops.com}.

\vspace{-1.8em}
\section{Adaptive Prefix-to-prefix Training Policy}
\label{sec:method}
\vspace{-0.6em}

Our idea is motivated by the way human interpreters translate an MU and ``wait'' policies. In this paper, our purpose is to split the streaming text into prefix pairs. Then, the prefix-to-prefix MT model is jointly trained on the original sentence pairs and pseudo prefix pairs. In the following, we first introduce our method to construct prefix pairs (Section~\ref{sec:cmtp}) and prefix pairs with \emph{m} future words (Section~\ref{sec:cmtpwkfw}). Finally, we describe the joint training details in Section~\ref{sec:jt}.
% Concretely, given a streaming text $s$, we incrementally detect whether the translation of a text clip $s_{\leq t}(t = 1,2,...)$ is a prefix of the full sentence’s translation, where $s_{\leq t}$ denotes the head $t$ words of $s$. Once a text clip $s_{\leq t}$ contains enough information to produces its translation that is a prefix of the full sentence’s translation $y^{t}$, $(x_{\leq t}, y^{t})$ is considered as an prefix pair.

\vspace{-1.4em}
\subsection{Constructing Prefix Training pairs}
\label{sec:cmtp}
\vspace{-0.6em}

Since there are no standard prefix pairs for training, we propose a simple method to automatically extract meaningful unit pairs to construct prefix training samples.

\begin{table}[!t]
    \centering
    \small
    % \renewcommand{\arraystretch}{1.4}
    % \resizebox{7.5cm}{!}{
    \resizebox{1.0\linewidth}{!}{
    \begin{tabular}{c|cc}
    \hline
    \multirow{2}{*}{Sentence Pair} & \multicolumn{2}{c}{\begin{CJK*}{UTF8}{gbsn} 牛顿 \ 发现 \ 了 \ 牛顿 \ 运动定律 \end{CJK*}} \\
    \cline{2-3}
    & \multicolumn{2}{c}{Newton discovered newton's laws of motion} \\
    \hline
    \multirow{2}{*}{Prefix Pairs} & \multicolumn{1}{c|}{\begin{CJK*}{UTF8}{gbsn} 牛顿 \end{CJK*}} & \multicolumn{1}{c}{\begin{CJK*}{UTF8}{gbsn} 牛顿 \ 发现 \ 了 \end{CJK*}} \\
    \cline{2-3}
    & \multicolumn{1}{c|}{Newton} & Newton discovered \\
    \hline
    \multirow{2}{*}{Prefix pairs with \emph{m} future words} & \multicolumn{1}{c|}{\begin{CJK*}{UTF8}{gbsn} 牛顿 \ [fw] \ 发现 \ 了 \end{CJK*}} & \begin{CJK*}{UTF8}{gbsn} 牛顿 \ 发现 \ 了 \ [fw] \ 牛顿 \ 运动定律 \end{CJK*} \\
    \cline{2-3}
    & \multicolumn{1}{c|}{Newton} & Newton discovered \\
    \hline
    \end{tabular}
    }
    \vspace{-0.8em}
    \caption{The training samples for the prefix-to-prefix MT model generated according to algorithm \ref{alg1} and its refined method, where \emph{m} is set to 2. ``[fw]'' is used to distinguish the source prefix from \emph{m} future words.}
    \label{table2}
    \vspace{-1.5em}
\end{table}

Before we describe our method, we first try to figure out what is an MU. MU is first proposed in \cite{zhang2020learning}, which is defined as \emph{the minimum segment whose translation will not be changed by subsequent text}. Therefore, we extend the definition of MU and then define an MU pair as \emph{the minimum segment pair which can translate from source to target}. We expect a source sentence prefix of a prefix pair contains enough information to produce a target sentence prefix like sentence pairs do (see Table~\ref{table1}). A simultaneous interpreter dynamically segments the sentence pair into 3 MU pairs and translates source MUs monotonically.

% Formally, we extract prefix pairs based on a pre-trained full sentence MT system $M_{nmt}$. Given a streaming source text $x = {x_{1}, x_{2}, ...x_{T}}$, we want to find a list of prefix pairs $S_{pref} = {S_{1}, S_{2}, ...S_{P}}$ i.e., to split $x$ into $P$ prefix pairs.

Our idea to generate parallel prefix pairs is that, for a source sentence prefix $\mathbf{x}_{\leq t} = {x_{1}, x_{2}, ...x_{t}}$, if its translation $\mathbf{y}^{t} = M_{nmt}(\mathbf{x}_{\leq t})$ is also a prefix of the candidates of full sentence's translation $\widetilde{\mathbf{y}} = M_{nmt}( \mathbf{x} )$, we add $(\mathbf{x}_{\leq t}, \mathbf{y}^{t})$ prefix pair into $\mathbf{S_{pref}}$. Otherwise, the model continues reading source words and judges whether the current source prefix is a prefix of the candidates of full sentence's translation. The reason is that $\mathbf{x}_{\leq t}$ contains enough information to produce a translation and $(\mathbf{x}_{\leq t}, \mathbf{y}^{t})$ is the minimum segment pair that can translate from source to target. To keep consistency between previous partial translations, we forcedly decode previous partial translations and then decode the following new sequence. The whole process is described in Algorithm \ref{alg1}. The algorithm reads source sentence word-by-word through a \emph{for} loop (Line~\ref{alg1:line3}), and generates translation by force decoding previous prefix translations, shown as $tgt_{forced}$ (Line~\ref{alg1:line4}). If its translation is a prefix of the candidates of full sentence's translation through beam search\footnote{In this paper, we use top 10 results as candidates.} (Line \ref{alg1:line5}), we add it into $\mathbf{S_{pref}}$ (Line~\ref{alg1:line6}).

\vspace{-1.4em}
\subsection{Constructing Prefix Training pairs With M Future Words}
\label{sec:cmtpwkfw}
\vspace{-0.6em}

The algorithm in section \ref{sec:cmtp} is intuitive and simple, however, it is based on an ideal READ policy, i.e., the ability to determine sufficient source prefix for translation, which is not always true in practice and results in a drop in translation accuracy and fluency. 
% For example, if we have a source sentence "\begin{CJK*}{UTF8}{gbsn} 今天 \ 是 \ 星期四 \ ？ \ 我 \ 以为 \ ... \end{CJK*}" and its translation "Is it Thursday ? i think  ...". We suppose the READ policy decides "\begin{CJK*}{UTF8}{gbsn} 今天 \ 是 \ 星期四 \end{CJK*}" is a source sentence prefix, then the WRITE policy in section \ref{sec:cmtp} produces its translation "It is Thursday" with high probability. Obviously, it is incorrect and can not be modified due to translation monotonicity. 

To alleviate it, we propose a refined method to extract prefix pairs with \emph{m} future words. Our method is inspired by ``wait'' policies for waiting a fixed number of source words to make source prefix more sufficient for translation (see Table~\ref{table2}).  The process is almost the same as algorithm \ref{alg1}, except that we modify the generation process of prefix pairs(Line \ref{alg1:line6}). Specifically, given a streaming source text $\mathbf{x} = \{x_{1}, x_{2}, ...\}$, if we detect a source prefix $\mathbf{x}_{\leq t}$ that contains enough information to produce its translation, we further add a special token ``[fw]'' and \emph{m} future words $f_{t} = \{x_{t + 1}, ..., x_{t + m} \}$ \footnote{\emph{m} is a hyper-parameter as the number of future words. Larger \emph{m} means higher latency. In this paper, we set \emph{m} = 2.} at the end of the source prefix. The reason is that adding $m$ future words acts as ``conjunctions'' to allow the model to take into account the fluency and articulation between partial translations, and provides more source information that leads to generating better translation at the same time. In addition, the refined method is the first to combine adaptive policy and fixed policy in WRITE policy to take advantage of their respective advantages, which achieves promising results against strong baselines and is also complementary to previous non-end-to-end studies. Take the sentence pair in Table~\ref{table1} as an example, we generate training samples as illustrated in Table~\ref{table2}. Note that for $t$ larger than $T - m$, we only use the remaining words in the source sentence as \emph{future} words. 
\begin{figure}[!t]
    \centering
    \includegraphics[width=0.8\columnwidth]{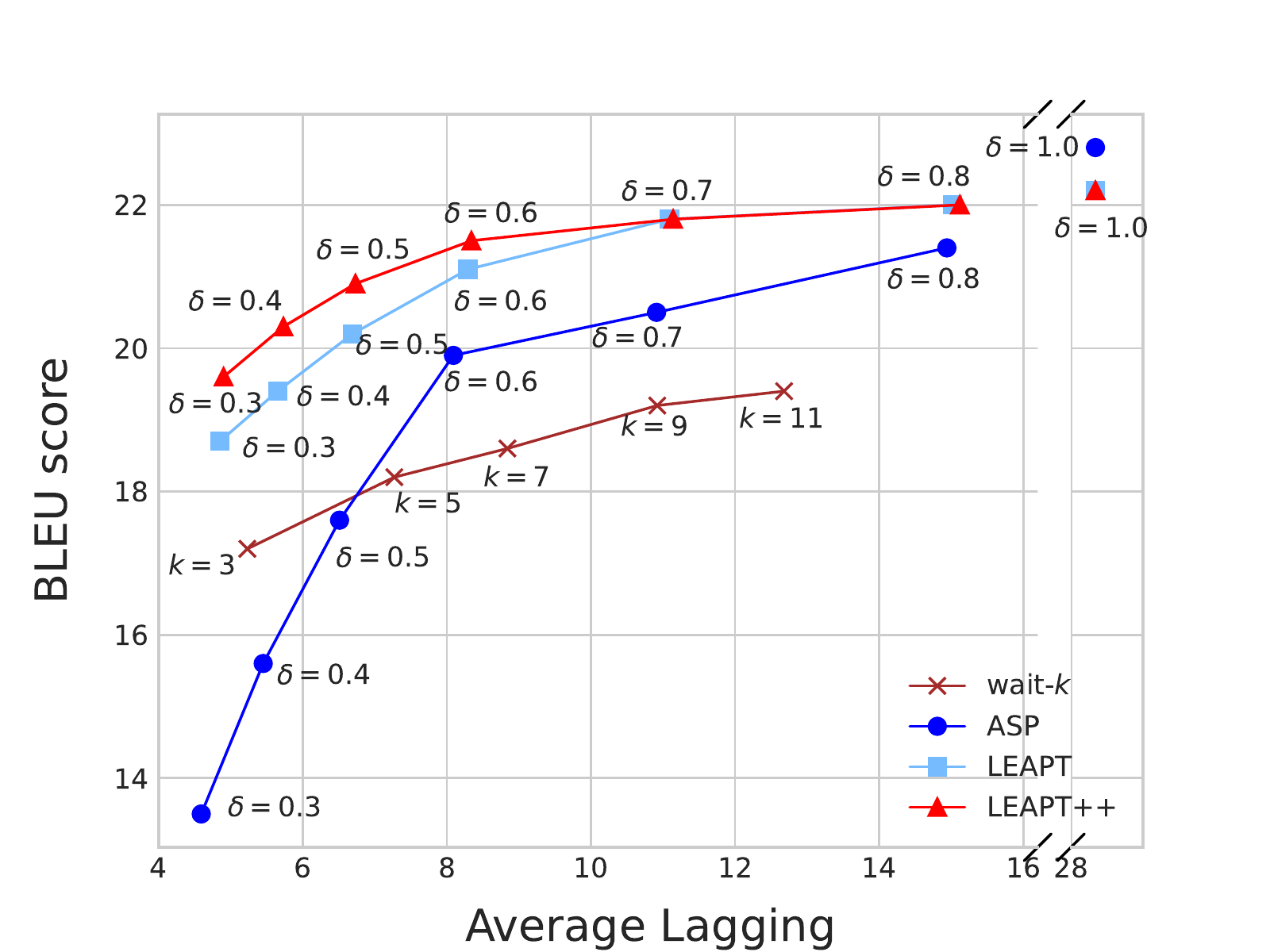}
    \vspace{-0.8em}
    \caption{Translation quality vs. latency results on the NJU-newstest-2017 dataset. Note that the full-sentence translation performance for each method is reported in upper right corner of the figure. $\delta$ is the threshold of the segmentation model of ASP.}
    \label{fig:fig2}
    \vspace{-1.4em}
\end{figure}

\begin{figure}[!t]
    \centering
    \includegraphics[width=0.8\columnwidth]{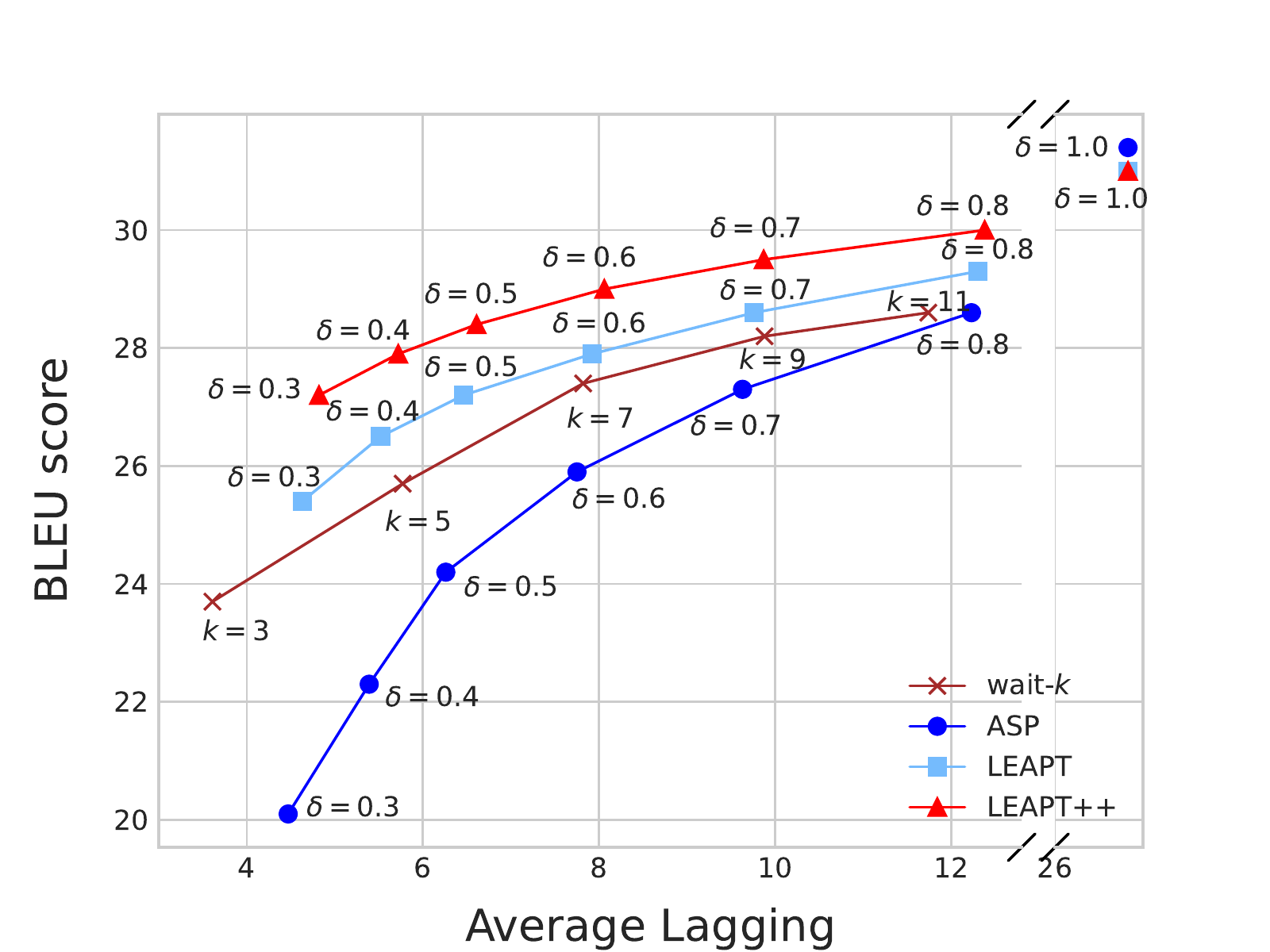}
    \vspace{-0.8em}
    \caption{Translation quality vs. latency results on the WMT15 German-English dataset.}
    \label{fig:fig3}
    \vspace{-1.2em}
\end{figure}

% Specifically, if we have a source sentence "\begin{CJK*}{UTF8}{gbsn} 今天 \ 是 \ 星期四 \ ？ \ 我 \ 以为 \ ... \end{CJK*}[Is it Thursday ? i think ...]" and the READ policy generates its prefix "\begin{CJK*}{UTF8}{gbsn} 今天 \ 是 \ 星期四\end{CJK*}", we further add a special token "[fw]" and \emph{m} future words \footnote{\emph{m} is a hyper-parameter as the number of future words. Larger \emph{m} means higher latency. In this paper, we set \emph{m} = 2.} at the end of the source prefix, i.e, "\begin{CJK*}{UTF8}{gbsn} 今天 \ 是 \ 星期四 \ [fw]  \ ？ \ 我 \end{CJK*}". Meanwhile, the target sentence prefix is the same as before, and then we obtain modified parallel prefix pairs $S_{pref}^{'}$. 

\vspace{-1.4em}
\subsection{Joint Training}
\label{sec:jt}
\vspace{-0.6em}

In the training stage, we extract the prefix pairs in the training corpus according to the basic method (Section~\ref{sec:cmtp}) or refined method (Section~\ref{sec:cmtpwkfw}). Specifically, for each source sentence $\mathbf{x} = \{x_{1}, x_{2}, ..., x_{T}\}$ in the training corpus, we generate $P$ prefix pairs. Each prefix pair consists of a source prefix and a target prefix, as illustrated in Table~\ref{table2}. Then, our prefix-to-prefix MT model is jointly trained on the original sentence pairs and pseudo prefix pairs.
\begin{algorithm}[!t]
        \footnotesize
	\SetAlgoLined
        \caption{Generating Parallel Prefix Pairs}\label{alg1}
	\KwIn{$\mathbf{x}=\{x_1, x_2, ..., x_T\}$}
        % \tcp*[f]{a source sentence} \\
	\KwOut{$\mathbf{S_{pref}}$}
        % \tcp*[f]{a set of parallel prefix pairs} \\
	$k=0$ \\
    % \tcp*{position of last MU boundary} 
	$\widetilde{\mathbf{y}}=M_{nmt}(src=\mathbf{x})$
    % \tcp*{$N$ best candidates via beam search}
	
	\For{$t=1,\dots,T$}{ \label{alg1:line3}
		$\mathbf{y}^t=M_{nmt}(src=\mathbf{x}_{\le t}, tgt_{forced}=\mathbf{y}^k)$ \label{alg1:line4} \\
        % \tcp*{forced decoding} \label{alg1:line4}
		\If{$\mathbf{y}^t\  is\  a\  prefix\  of\  \widetilde{\mathbf{y}}$}{ \label{alg1:line5}
			$\mathbf{S_{pref}} = \mathbf{S_{pref}} \cup (\mathbf{x}_{\le t}, \mathbf{y}^t)$ \label{alg1:line6} \\
			$k=t$
		}
	} 
	\textbf{return} $\mathbf{S_{pref}}$
\end{algorithm}

As mentioned in Section \ref{sec:intro}, simultaneous machine translation involves a READ policy and WRITE policy, where a READ policy and WRITE policy are combined to achieve high translation quality and an as-short-as-possible delay in an e2e or ne2e way. 
% However, for WRITE policies of non-end-to-end approaches, they do not consider the input mismatch between training and inference, which results in a drop in translation accuracy and consistency between partial prefix translations. To alleviate it, we propose an adaptive prefix-to-prefix training policy, which can generate higher-quality translation with corresponding latency and is also complementary to previous non-end-to-end studies. 
In order to fully demonstrate the effectiveness of our methods, we choose one of the READ policies of ne2e adaptive approaches to combine with our proposed WRITE policies to do simultaneous interpretation. The reason is that the adaptive methods are more flexible than the fixed ones and achieve better results. Concretely, we use the ASP method as our READ policy and only replace the WRITE policy of ASP with our proposed methods, and keep the rest intact. ASP is easy to implement and one
of recent work that performs competitive performance on
adaptive policies while others are not open source~\cite{lin-automatic-2022}.
% referred to~\cite{zhang2020learning} for details.

% All printed material, including text, illustrations, and charts, must be kept
% within a print area of 7 inches (178 mm) wide by 9 inches (229 mm) high. Do
% not write or print anything outside the print area. The top margin must be 1
% inch (25 mm), except for the title page, and the left margin must be 0.75 inch
% (19 mm).  All {\it text} must be in a two-column format. Columns are to be 3.39
% inches (86 mm) wide, with a 0.24 inch (6 mm) space between them. Text must be
% fully justified.

\vspace{-1.6em}
\section{Experiments}
\label{sec:exp}
\vspace{-0.6em}

\begin{table*}[!t]
    \centering
    \footnotesize
    % \small
    \renewcommand{\arraystretch}{1.6}
    \resizebox{17.5cm}{!}{
    % \resizebox{1.0\linewidth}{!}{
        \begin{tabular}{c|c}
        \hline
        Source & \begin{CJK*}{UTF8}{gbsn} 环境 \ 问题 \ 表现 \ 在 \ 大气 \ 、 \ 水 \ 、 \ 土壤污染 \ 上 \ ， \ 但 \ 根子 \ 还是 \ 在 \ 生产 \ 生活 \ 方式 \ 上 \end{CJK*} \\
        \hline
        ASP & \textcolor{red}{WAIT*6} the expression of environmental problems \textcolor{red}{WAIT*2} in the atmosphere , \textcolor{red}{WAIT} water \textcolor{red}{WAIT} , \textcolor{red}{WAIT} and soil pollution \textcolor{red}{WAIT*2} , \textcolor{red}{WAIT} he said . \textcolor{red}{WAIT*2} the roots are still , \textcolor{red}{WAIT*3} the way of life in production . \\
        \hline
        LEAPT & \multicolumn{1}{l}{\textcolor{red}{WAIT*6} environmental problems show in \textcolor{red}{WAIT*2} the atmosphere , \textcolor{red}{WAIT} water \textcolor{red}{WAIT} , \textcolor{red}{WAIT} soil pollution \textcolor{red}{WAIT*2} , \textcolor{red}{WAIT} he said . \textcolor{red}{WAIT*2} the roots are still \textcolor{red}{WAIT*3} in the production lifestyle .} \\
        \hline
        LEAPT++ & \multicolumn{1}{l}{\textcolor{red}{WAIT*6} environmental problems appear in \textcolor{red}{WAIT*2} the atmosphere , \textcolor{red}{WAIT} water \textcolor{red}{WAIT} , \textcolor{red}{WAIT} and soil pollution \textcolor{red}{WAIT*2} , \textcolor{red}{WAIT} but \textcolor{red}{WAIT*2} roots are still \textcolor{red}{WAIT*3} in the production lifestyle .} \\
        \hline
        \end{tabular}
    }
    \vspace{-0.8em}
    \caption{A source sentence and its translation with ASP, LEAPT and LEAPT++. \textcolor{red}{WAIT*i} denotes waiting for \emph{i} source words.}
    \label{table3}
    \vspace{-1.8em}
\end{table*}

We evaluate our proposed methods on two translation tasks: the NEU2017 Chinese-English (Zh-En) translation task (2M sentence pairs) \footnote{http://mteval.cipsc.org.cn:81/agreement/description}, and the WMT 2015 German-English (De-En) translation task (4.5M sentence pairs) \footnote{http://www.statmt.org/wmt15/translation-task.html}. We use case-sensitive detokenized BLEU\cite{papineni2002bleu} score and Average Lagging (AL) \cite{ma2018stacl} to evaluate translation quality and latency respectively. $\delta$ is the threshold for the segmentation model of ASP.

\vspace{-1.4em}
\subsection{Data Settings}
\vspace{-0.6em}

We use an Chinese Tokenizer\footnote{https://github.com/fxsjy/jieba} to preprocess Chinese text and Moses Tokenizer\footnote{https://github.com/moses- \\ smt/mosesdecoder/blob/master/scripts/tokenizer/tokenizer.perl} to preprocess German and English text. All text is encoded by byte-pair encoding~\cite{sennrich2015neural} (BPE) to reduce the vocabulary sizes. For Zh-En, we use NJU-newsdev-2017 as our dev set and NJU-newstest-2017 as our test set, with 1,894 and 1,000 sentence pairs respectively. For De-En, we use newstest 2013 (3,000 sentence pairs) and newstest 2015 (2,169 sentence pairs) as our validation and test set.

\vspace{-1.4em}
\subsection{Model Settings}
\vspace{-0.6em}

We compare our methods with previous strong SiMT systems,
% (1) wait-k~\cite{ma2018stacl}: first waits for \emph{k} source words, and then translates a target word after reading each source word (fixed and end-to-end method); (2) ASP~\cite{zhang2020learning}: first detects whether the current source prefix is a complete meaning unit, and then feeds it into a full-sentence MT model for translation until generating end of sentence token, where we use the basic method for comparison (adaptive and non-end-to-end method); (3) LEAPT: first judges whether to wait for another source word via the basic READ policy of ASP, and then feeds it into our basic method for translation; (4) LEAPT++: first judges whether to wait for another source word via the basic READ policy of ASP, and then feeds it into our refined method for translation.
which contain the most widely used policies of fixed and adaptive methods.
\begin{itemize}
    \vspace{-0.7em}
    \item \emph{wait-k}~\cite{ma2018stacl}: first waits for \emph{k} source words, and then translates a target word after reading each source word (fixed and end-to-end method).
    \vspace{-0.7em}
    \item \emph{ASP}~\cite{zhang2020learning}: first detects whether the current source prefix is a complete meaning unit, and then feeds it into a full-sentence MT model for translation until generating end of sentence token, where we use the basic method for comparison (adaptive and non-end-to-end method).
    \vspace{-0.7em}
    \item \emph{LEAPT}: first judges whether to wait for another source word via the basic READ policy of ASP, and then feeds it into our basic method for translation.
    \vspace{-0.7em}
    \item \emph{LEAPT++}: first judges whether to wait for another source word via the basic READ policy of ASP, and then feeds it into our refined method for translation.
\end{itemize}
\vspace{-0.7em}

% For strong baselines, we use the same implementations for wait-\emph{k} and ASP, referred to \cite{ma2018stacl} and \cite{zhang2020learning} for details.
For the READ policy, we use the same implementation as~\cite{zhang2020learning} and only replace the full-sentence MT model with our proposed methods to fully demonstrate the effectiveness of our methods. Specifically, the segmentation model is based on the classification task of BERT~\cite{devlin2018bert} (Chinese\footnote{https://huggingface.co/bert-base-chinese} and German\footnote{https://huggingface.co/bert-base-german-cased} version), which determines whether to wait for another source word through its prediction (1 or 0)\footnote{If the prediction is 0, the segmentation model needs to read another source word, otherwise the current source prefix can be fed into the WRITE policy for translation.}. The training process of the segmentation model is based on BERT base version and following fine-tuning process~\cite{devlin2018bert}.

For the WRITE policy, we use the base Transformer~\cite{vaswani2017attention} for pre-training and fine-tuning with the learning rate of $3e^{-4}$, $3e^{-5}$ respectively. Besides, the prefix-to-prefix MT model uses beam search with a beam size of 10 during decoding.

\vspace{-0.6em}
\subsection{Overall Results}
\vspace{-0.6em}

Figure~\ref{fig:fig2} and~\ref{fig:fig3} show the quality-latency results on NJU-newstest-2017 dataset and newstest 2015 dataset respectively. We observed that:
\begin{itemize}
    \vspace{-0.7em}
    \item Our methods outperform wait-\emph{k} and ASP methods in terms of translation accuracy and latency for Zh-En and De-En experiments.
    \vspace{-0.4em}
    \item Our methods perform more robustly than wait-\emph{k} and ASP methods in both low and high latency for Zh-En and De-En experiments. In particular, our methods generate higher-quality translations in the same low latency with respect to the strong baselines.
    \vspace{-0.6em}
    \item As $\delta$ increases, the quality and latency also increase accordingly. Therefore, $\delta$ can be tuned to obtain a balance between quality and latency according to different requirements.
    \vspace{-0.6em}
    \item Compared to LEAPT, LEAPT++ significantly improve the translation quality with a slight increase in latency, which fully proves our previous hypotheses.
\end{itemize}
\vspace{-0.6em}

% Surprisingly, wait-\emph{k} performs slightly better than ASP in low latency for the Zh-En experiment while outperforming ASP in terms of latency for the De-En experiment, which is somewhat contradictory to the findings of ASP. The reasons we guess are different datasets and implementations for ASP. Due to its code is not open-source, we reproduce the ASP method based on \cite{zhang2020learning}, which is likely to cause the inconsistency between our results and those reported by ASP. 
Note that the prefix-to-prefix MT model is slightly worse than the full-sentence MT model when translating full sentences, which is within our expectations.

\vspace{-1.2em}
\subsection{Case Study}
\vspace{-0.6em}

Due to length limitations, we only provide an example of ASP vs. our proposed methods in Table~\ref{table3}, where $\delta$ is set to 0.4. It is clear that our proposed methods are capable of generating higher-quality translations when translating source prefixes. For example, given ``\begin{CJK*}{UTF8}{gbsn}环境 \ 问题 \ 表现 \ 在 \ 大气 \ 、\end{CJK*}'' , ASP generates ``\textbf{the expression of environmental problem}'' and our methods generate ``\textbf{environmental problems show in}'' and ``\textbf{environmental problems appear in}'' respectively, which translate ``\begin{CJK*}{UTF8}{gbsn}表现 \ 在\end{CJK*}'' correctly while ASP ignores it. Compared to LEAPT, LEAPT++ generates not only more accurate but also coherent translations. The most straightforward illustration in the example (see Table~\ref{table3}) is that LEAPT produces ``\textbf{he said.}'' while LEAPT++ produces ``\textbf{but}'', which confirms our above hypotheses of adding \emph{m} future words as ``conjunctions''.

\vspace{-1.2em}
\section{Conclusion}
\vspace{-1.0em}
In this paper, we propose a novel adaptive prefix-to-prefix training policy for SiMT, referred to as \textbf{LEAPT}. Motivated by strategies utilized by human interpreter and ``wait'' policies, the proposed prefix-to-prefix MT model is a generic and better WRITE policy for non-end-to-end approaches. Compared to strong baselines, our proposed methods greatly improve the translation quality in both low and high latency and are complementary to previous ne2e studies.

\bibliographystyle{IEEEbib}
\bibliography{refs}

\end{document}